Outer Loop     Inner Loop     Experts Interacting

# BADGER: Learning to (Learn [Learning Algorithms] through Multi-Agent Communication)

Marek Rosa[1], Olga Afanasjeva[1], Simon Andersson[1], Joseph Davidson[1], Nicholas Guttenberg[1,2,3],
Petr Hlubuček[1], Martin Poliak[1], Jaroslav Vítku[1] and Jan Feyereisl[1]

info@goodai.com

## Abstract

In this work, we propose a novel memory-based multi-agent meta-learning architecture and learning procedure that allows for learning of a shared communication policy that enables the emergence of rapid adaptation to new and unseen environments by learning to learn learning algorithms through communication. Behavior, adaptation and learning to adapt emerges from the interactions of homogeneous experts inside a single agent. The proposed architecture should allow for generalization beyond the level seen in existing methods, in part due to the use of a single policy shared by all experts within the agent as well as the inherent modularity of 'Badger'.

***Disclaimer***: *Our aim of releasing this technical report about our preliminary and ongoing work is to start a discussion with others interested in the discussed and related topics, for recruiting purposes, and to inform anyone interested about what we are working on.*

## Motivation

A complex adaptive system can be viewed as a multi-agent system where many agents form networks, communicate, coordinate, update their state, adapt, and achieve some shared goals (Holland 2015; Waldrop 1992; Lansing 2003).

Many complex living forms, including humans and their brains can also be described as a multi-agent system. In the brain, for example, biological neurons being the agents, and the synapses and neurotransmitters the communication channels (Solé, Moses, and Forrest 2019; Sporns 2010; Avena-Koenigsberger, Misic, and Sporns 2017).

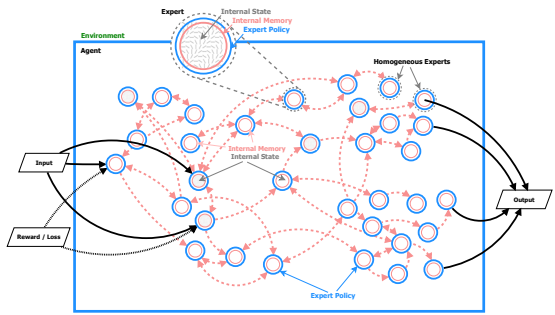

**Figure 1.** Illustration of a 'Badger' agent. A single agent comprises a number of experts (○) that operate according to the same fixed and shared policy (○). Each expert has its own unique internal state (○). Communication (---•), resulting from the execution of the fixed shared policy, with varying inputs (⌒•) per each expert (i.e. incoming messages and expert's internal state), gives rise to learning algorithms able to solve new and unseen tasks.

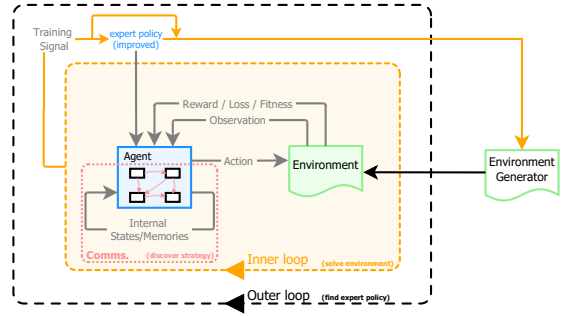

**Figure 2.** Overview of the inner and outer-loop learning procedure discussed in this work. Unlike in meta-reinforcement learning, a third stage occurs inside the agent where experts communicating with each other give rise to learning algorithms adapting to novel tasks. Figure inspired by (Botvinick et al. 2019)

## SUMMARY

An architecture and a learning procedure where:

- An **agent** is made up of **many experts**
- All experts **share the same communication policy (expert policy)**, but have **different internal memory states**
- There are **two levels of learning**, an inner loop (with a **communication stage**) and an outer loop
- **Inner loop** – Agent's behavior and adaptation should emerge as a result of experts communicating between each other. Experts send messages (of any complexity) to each other and update their internal states based on observations/messages and their internal state from the previous time-step. Expert policy is fixed and does not change during the inner loop
- **Inner loop loss** need not even be a proper loss function. It can be any kind of structured feedback guiding the adaptation during the agent's lifetime
- **Outer loop** – An expert policy is discovered over generations of agents, ensuring that strategies that find solutions to problems in diverse environments can quickly emerge in the inner loop
- **Agent's objective** is to adapt fast to novel tasks

Exhibiting the following novel properties:

- **Roles** of experts and **connectivity** among them assigned **dynamically** at inference time
- Learned communication protocol with **context dependent messages** of varied complexity
- **Generalizes** to different numbers and types of inputs/outputs
- Can be trained to handle **variations in architecture** during both training and testing

Initial empirical results show **generalization** and **scalability** along the spectrum of learning types.

However, while there are models of neuron and synapse dynamics, we are still discovering new things about the 'policies' of biological neurons and the ways in which they communicate with each other. Similarly, yet at the other end of the scale of complex adaptive systems, when looking at societies and learning within them (Bandura and Walters 1977), we know it is not only one individual who is born

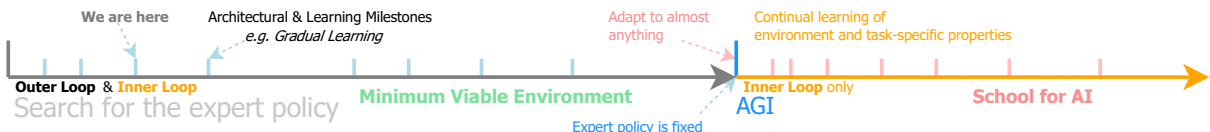

**Figure 3.** Roadmap of 'Badger', depicting the development stage (search for the expert policy) and the deployment stage (expert policy is fixed, inner loop execution only). Once the expert policy is found, it can stay fixed, and further learning is performed only by experts communicating with each other while updating their internal memory states. The expert policy will not change anymore. The expert policy should foster the emergence of agent's properties such as: incremental learning, continuous learning, transfer learning, lifelong learning, intrinsic motivation for experts, intrinsic motivation for the agent, etc. We are aiming to learn an expert policy, which when integrated into a dynamical system made of many experts, should demonstrate complex self-organizing global behavior, emerging from local interactions among experts.

[1]GoodAI, [2]Cross Labs, [3]Earth-Life Science Institute, Tokyo Institute of Technology

## GLOSSARY

**Agent -** A decision-making unit interfacing with the world/environment. Comprises of *multiple experts*.

**Expert -** A decision-making sub-unit of an agent. Comprises of an *expert policy* (same across all experts) and an *internal memory/state* (unique to each expert). Collectively, *via communication*, experts should give rise to learning algorithms.

**Expert Policy –** a function defining a strategy for *communicating among experts* within an agent. *Same* for all experts within an agent. Functionally it captures the notion of *'how can experts coordinate together to learn to solve a new task/environment as fast as possible'.*

**Outer Loop –** The search for the expert policy. Represents g*enerations* of agent's behavior over which the expert policy is discovered/learned. Agent is trained over multiple environments/tasks. The expert policy is learned by *adjusting the weights* of a model (e.g. a neural network) that represents the expert policy.

**Inner Loop –** Behavior of an agent during its *lifetime*. Parameters of the model representing the expert policy are *fixed* and are not changed during the inner loop. Each inner loop step involves a communication stage. Inner loop loss need not even be a proper loss function, any useful feedback might suffice.

**Communication –** At each step of the inner loop, experts can *send messages* to each other and *update their internal states* based on observations/messages and their internal state from the previous time-step. The exchange of messages can happen more than once per inner loop step.

with the ability to solve every task, but rather individuals have the flexibility to be trained as specialists using information that is distributed throughout the society as a whole.

In this work, we are interested in collective decision making at different scales, yet within a single agent. We use the word 'experts' as a name for agents inside an agent, as depicted in *Figure 1*. This way, there's no confusion whether we are talking about agents in an environment, or experts inside an agent.

The 'Badger' architecture aims to go further than this, it strives to automate the search for an expert policy, by framing the problem as multi-agent learning (in our case, multi-expert learning). We are searching for one, universal expert policy, used in all experts (although, we expect it may be useful to have more than a single expert policy, yet fewer policies than experts). This process can be seen in *Figure 2* on the previous page.

If trained on environments that select for specific agent properties, we expect that we can search for an expert policy, from which agent properties can emerge, such as incremental learning, online learning, overcoming forgetting, gradual learning, recursive self-improvement, and many more.

We expect the expert policy to be fixed for the duration of an agent's life, the inner loop, therefore search for it happens in the outer loop. This means that any learning inside an agent should be a result of communication among experts, and changes of their internal states.

Conceptually, this process can be seen in *Figure 4*, where, in order to solve problems, experts communicate to find and employ a suitable strategy. This can only possible after an expert policy is found that enables such collective behavior.

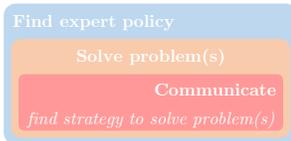

**Figure 4.** Conceptual view of the nested two-level learning procedure with multi-agent communication.

Since experts learn to determine on their own who should communicate with whom, the overall processing of the agent can be made decentralized, meaning that in principle new instances of experts can be asynchronously added or removed.

## Architecture

In this framework, an agent is made up of multiple experts. All experts share one homogeneous expert policy (henceforth, expert policy, see Glossary).

An expert policy can be implemented as a neural network, a handcrafted program or in other ways. Experts in an agent can share more than a single expert policy, but it is expected that the total number of policies should be much smaller than the number of experts. A network made from experts that have a shared neural network policy and a handcrafted policy is also possible.

Each expert is equipped with a unique and distinct internal memory. Internal memory of an expert holds information about the expert's own internal state (e.g. neural network activations, hidden state of an RNN, program variables, etc.). Adaptation to novel environments (learning how to solve tasks within them), as well as adaptation of this adaptation (learning how to learn to solve novel tasks/environments), should be the result of communication between experts and changes of their internal states. Agents should learn how to learn to solve new and unseen tasks/environments rapidly.

Once an expert policy is trained (through the outer loop), an agent doesn't learn through changes to the expert policy (e.g. no changes of weights in a neural network) – the expert policy is fixed during the agent's lifetime (the inner loop). Agent should learn only via experts communicating and through updates to their internal states, in order to solve, or learn to solve new tasks or new environments.

**Topologies**

The configuration of experts within an agent, in the 'topographical' sense, and the determination of who interacts with whom is of vital importance as it governs the internal dynamics of communication and hence the emergence of learning algorithms.

Our starting assumption is that every expert can communicate with every other expert. Naturally, as the network grows, this would quickly become infeasible.

The ultimate goal is entirely learned and dynamic topology where, rather than fixed connectivity, experts determine whom to talk to at inference time, depending on the task/environment that is being solved, internal states of experts and messages that they receive. Stepping away from fixed and rigid topology has recently been shown to yield intriguing results (Ha, David, Andrew Dai 2016; Gaier and Ha 2019).

We have experimented with the following approaches:

- *Hardwired topologies* where input experts propagate signals to hidden layer experts with recurrent connections, followed by propagation to output experts.

- *Hardwired* and *'Homogeneous' topologies* - no difference between input, hidden and output layers

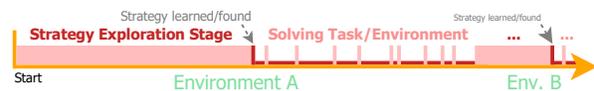

**Figure 5.** Temporal depiction of the **inner loop**. Before an agent can solve a new task/environment, it will need to undergo an **exploration stage** through which structures, patterns or other types of identifiers can be discovered to help identify and help discover the strategy to use to solve the presented task/environment. This, most likely unsupervised stage, will require the existence of exploration meta-strategies such as novelty search, affordances or other methods that will enable the discovery of novel algorithms. Unlike depicted above, there might not be a clear boundary among tasks/environments in the real world.

- *Dynamic topologies* - e.g. via attention module of the Transformer architecture (Vaswani et al. 2017)

- *Random topologies* – enforce the expert policy to be as much invariant to agent topology as possible

The above are examples of some of the approaches we have tried thus far, yet many other methods and combinations are possible, each with different benefits and drawbacks, for example, in terms of scalability, ease of training or representational capacity.

## Training Procedure

In the proposed framework, training is performed using a two-stage training procedure. This procedure comprises of an outer loop and an inner loop. This falls within the areas of meta-learning (Bateson 1972; Jurgen Schmidhuber 1987; Bengio 2000; Hochreiter, Younger, and Conwell 2001; Grefenstette et al. 2019) and bi-level optimization (Colson, Marcotte, and Savard 2007; Sinha, Malo, and Deb 2018; Franceschi et al. 2018). As, in our case, learning and adaptation happens in the internal states of experts. This can also be viewed as a form of memory-based meta-learning, c.f. (Ortega et al. 2019).

### Outer Loop

The outer loop corresponds to the search for the expert policy. The agent is trained across an array of different environments. The expert policy can be parameterized via a neural network or other models, whose parameters, also called meta-parameters (Grefenstette et al. 2019), are optimized to maximize performance on environments both seen and unseen by the inner loop. The expert policy found should be general to a large set of seen and unseen environments and tasks. The outer loop training can also be viewed as a multi-agent (e.g. reinforcement) learning problem (Tuyls and Weiss 2012).

*Manual/Handcrafted Expert Policy*: A parallel approach to searching for the expert policy via the outer loop is to 'just program it', like a regular hand coded program. We are investigating this handcrafted expert policy approach, because any progress in it can lead to adding more useful constraints to the outer loop search, which might lead to its improvements. However, this topic is out of the scope of this paper. We will not go into details, except saying that we believe that the minimum requirements for a handcrafted expert policy could be: *experts detecting and generating patterns, simple credit assignment, modifiable experts after receiving special messages from other experts.*

### Inner Loop

The inner loop should correspond to behavior during an agent's life time. The expert policy is now fixed; the weights/meta-parameters are not adjusted throughout. The agent is presented with a single environment or a set/sequence of environments and tasks and experts within an agent should begin to communicate with each other based on incoming data from presented environment(s). The agent should quickly adapt by experts communicating with each other and by changing their internal states. Changes in the internal states of experts should give rise to an algorithm that enables adaptation of an agent to quickly learn to solve the new environment/task it is presented with. As depicted in *Figure 5*, it is expected that before an agent can solve a new task/environment, it needs to undergo an exploration stage to help identify and discover the strategy to use to solve the presented task/environment. This, most likely unsupervised stage, will likely require the knowledge of exploration meta-strategies such as novelty search (Stanley 2019), affordances or other methods that will enable the discovery of novel algorithms.

*Communication Stage*: At each step of the inner loop, experts can interact with each other by exchanging messages one or more times, until a predefined or learned condition is reached, e.g. an agreement among experts. The types of messages and the communication protocol are all learned and **can vary greatly**, depending on the learned expert policy and the context.

**Loss functions**
There are different loss functions for each of the two training loops, i.e. the outer loop loss function and the inner loop loss function (one or many). The agent might receive the inner loop loss on the input inside the inner loop, in order to understand what has to be minimized during the inner loop. In fact, the inner loop loss need not even be a proper loss function per se, but could be any kind of structured feedback so long as it relates eventually to the outer loop performance. Examples of this can be a reward, a supervision signal, explicit gradients, future outcomes given proposed actions, or almost anything that is informative to the outer loop. The experts can just treat it as another informative input. The expert policy is trained by adjusting its (meta-)parameters to minimize the outer loop loss function. The outer loop loss function is designed to reward rapid adaptation to new environments and rapidly learning to adapt to new environment/task families. The structure of the outer loop loss function should give rise to learned communication behavior in the inner loop of the training procedure. As mentioned in the previous section, the communication stage might also benefit from an explicit loss or objective function, either fixed or learned.

**On the difficulty of training learned learners**
Due to the fact that the proposed Badger architecture comprises of multiple loops of optimization, an inherent issue with systems of this types exists (Metz et al. 2019). The inner loop, especially when containing many steps and, in addition, the communication stage upon which Badger depends, result in many steps of optimization through which gradients or other relevant information for learning needs to propagate. To alleviate such problems may require advances in optimization (Beatson and Adams 2019), the use of optimization methods such as evolutionary methods which do not need to propagate signals to perform credit assignment (Maheswaranathan et al. 2019) or techniques such as initialization strategies to improve the quality of received gradients.

**Expert Uniqueness and Diversity**
The fact that a single policy is used by all experts poses one clear challenge right from the start. The issue of diversity, or rather lack thereof, also sometimes called 'module collapse' (Shazeer et al. 2017). Assuming that the internal states of all experts are initialized identically and all experts receive the same input, then all experts would behave identically, hence no interesting behavior would ever emerge. For this reason, one important aspect of research is how to enforce diversity among experts during training. Some possible approaches that have already been tested by us include unique initial random internal states, ensuring different experts receive different input, explicit identifiers supplied on input to each expert, to explicitly force differentiation, or regularization. There are, however, other methods that warrant exploration in the future, e.g. (Cases et al. 2019).

## Scalability and Growing

We believe, that the modularity of Badger and homogeneity of the expert policy should not only allow for better generalization (Chang et al. 2018; Rosenbaum et al. 2019) but also for greater scalability. Irrespective of the number of experts, only one expert policy is trained. The architecture can be grown without re-training. Adding an expert to an agent is performed trivially by simply copying or cloning an expert or instantiating a new expert.

Task-specific algorithms are expected to emerge in expert's internal states, as well as, as a result of the dynamics of the communication between the experts. Hence there is no need to change the meta-parameters, i.e. the model/policy weights. In standard neural networks, such change would invalidate the trained model. Adding more experts should allow for greater computational and storage capacity and increased ability to learn more diverse set of learning algorithms.

An example learning procedure that could show the growing ability:

1. Train an agent via the outer loop
   a. i.e. learn an expert policy that is general across many environments, possibly via a curriculum
2. Fix expert policy - weights cannot be changed anymore
3. Run agent in a new environment
   a. Agent rapidly adapts to learning to solve tasks in new environment through inter-expert communication
   b. Emergence of task-specific algorithm/solution/policy in the communication dynamics of experts and in internal states of experts within the agent
4. Add more experts to agent by cloning experts
   a. Feasible due to homogeneity of expert policy
   b. More computational/learning/adaptation capacity is obtained without re-training via the outer loop
5. Emergent algorithm present in expert's internal states can benefit from additional experts by offloading learning/computation to added experts, agreed upon via communication

## Generality

Badger architecture's focus is on learning how to learn fast and on learning an expert policy that is general in the sense that it is applicable to as many different environments as possible.

Unlike a policy in reinforcement learning (Sutton and Barto 2018), an expert policy in Badger ought to be invariant to the task and environment at hand, more like a policy in meta-reinforcement learning (Jürgen Schmidhuber 1995; J. X. Wang et al. 2016; Duan et al. 2017). An expert policy should encode a general communication strategy, a meta-strategy, that would allow decision-making subunits (experts) to communicate with each other and collectively derive task-specific strategies, based on incoming observations. We believe, such meta-strategy should allow generalization to classes of unseen problems rather than only to instances of unseen data from a single environment or task.

The characteristics of a Badger agent can then be shaped by exposing it to different environments:

- Environments can select for desired learning or adaptability traits or provide challenges of increasing difficulty (curriculum learning).

- An example of a desired learnable trait is gradual learning (need for growth of experts)

- Training can be focused by providing the simplest possible environment that selects for a desired capability (minimum viable environment).

- Experts in a Badger agent can also learn to perform automatically the division of labor, necessary to solve new tasks and automatic allocation (by themselves/or by others) to parts of the solution space that requires their attention or processing capacity

### Dimensions of Generalization

One of the primary goals of machine learning and artificial intelligence is the development of algorithms and associated models that have strong generalization capabilities. This means that when a system is trained on some data, it is able to transfer knowledge obtained during training to new and unseen observations during testing. Meta-learning takes this a step further and rather than only being able to generalize to new observations, it is able to transfer and exploit knowledge onto new, but related distributions of tasks.

We believe that our Badger architecture can take generalization even further. It is not only concerned with being able to transfer knowledge to new observations or classes of tasks, but also to different types of learning problems.

Frequently, different types of learning are discussed in a way that invokes discreteness in the types of learning problems that exist. On the contrary, learning problems can be viewed as a spectrum or even a continuum, where discreteness and existing categorization is a concept frequently used to simplify dealing with a particular class of related problems.

We believe that Badger will allow for generalization to new and unseen types of learning problems. For example, if we train our system on optimization problems, unsupervised and supervised learning tasks, Badger should be able to generalize to bandit or even reinforcement learning types of problems, without encountering such problems during training.

We found, for example, that when we tried to train a Badger architecture on an identity function (that is, to output values provided on its inputs) in the presence of a hot/cold signal, it instead relied upon the hot/cold signal to learn a more general low dimensional convex function optimization strategy akin to triangulation.

## Recursive Self-improvement

Since Badger experts share a policy, any task specialization which they perform must be encoded in their internal state. As such, a single Badger expert should be able to learn to encode, coordinate, and deploy a number of different behavioral strategies. If this generalizes sufficiently to represent even 'potential' strategies - that is, things which would be primarily beneficial for a task which lies outside of the training set, then this suggests the potential for recursive self-improvement (Jurgen Schmidhuber 1987). The distinction between the case of Badger architectures and other forms of meta-learning is the weights representing the expert policy comprise only a fraction of the capacity of that which is provided by the activations of a large group of experts - as such, while there is less capacity for particular task-specific strategies to be memorized, there is at the same time still a potentially large and extensible capacity available to the agent overall.

Specially crafted loss functions used during the two-stage training procedure might specifically pressure the creation of an expert policy that might possess the necessary traits of recursive self-improvement, e.g. when experts self-organize for faster adaptation of the adaptation procedures themselves.

## Minimum Viable Environments, Autocurricula and Self-Play

Until now, we haven't described what kind of tasks are needed to guide the learning during the outer loop to learn a general enough expert policy.

We are proposing to create a minimum viable environment (MVE), which can be handcrafted, or with characteristics of autocurricula/self-play (Leibo et al. 2019; R. Wang et al. 2019), but whose properties and agent evaluation criteria, will promote learning of an expert policy that becomes increasingly more general.

### Why an MVE?

Training an agent in the real world, or in a real-world-like environment is extremely inefficient. Instead, if feasible, it is highly desirable to only focus on and implement an environment with minimal requirements that can then generalize to a human-relevant real-world environment.

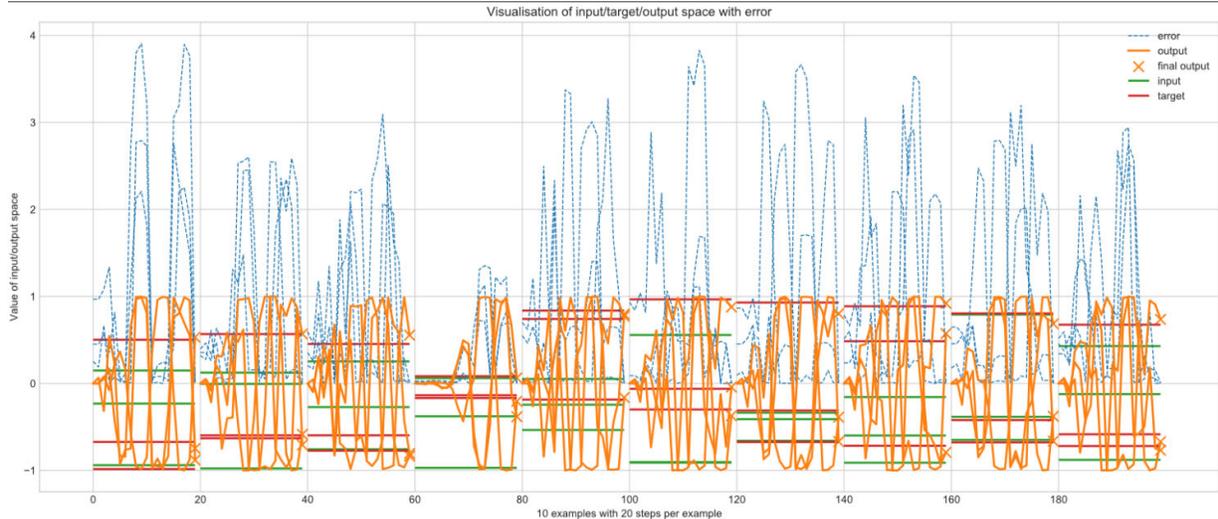

***Figure 6.*** This figure shows 10 different rollouts of a trained Badger expert solving the 'guessing game' task. Orange lines depict the output of the agent, while green and red lines show the input and target values, correspondingly. It is apparent that the expert policy discovers a strategy that oscillates the agent's output until it 'hits' the correct target at step 20 of the inner loop. Dashed blue line shows the communication values passed between the three experts making up the Badger expert. Only one expert receives the error information and hence the expert policy needs to possess the ability to communicate this information to the other two experts.

The motivation for an MVE is the observation that human intelligence is able to solve a wide array of problems which were not present under the conditions in which it evolved. Therefore, in some sense, the environment necessary for evolution to discover very general methods of learning did not need to encompass nearly as high a degree of complexity as the resulting methods were able to ultimately deal with.

**What do we believe were the basic requirements?**
Historically, humans had to frequently imagine things that didn't exist. We had to communicate in large groups, negotiate, have empathy, and so on. The human hand has five fingers and an opposable thumb, with enough dexterity that allowed the creation of tools (Perez 2018). On top of this, sexual selection (Darwin and Wallace 1858; Miller 2011) also guided the process, not just 'survival of the fittest'. All this enabled the evolution of the human intelligence as we know it, and that can be used to solve tasks that were not present in the original MVE (e.g. programming, directing a movie, discovering science)

*In other words, an MVE should be as simple as possible, while allowing the evolution/learning of human level intelligence and generality, but not simpler.*

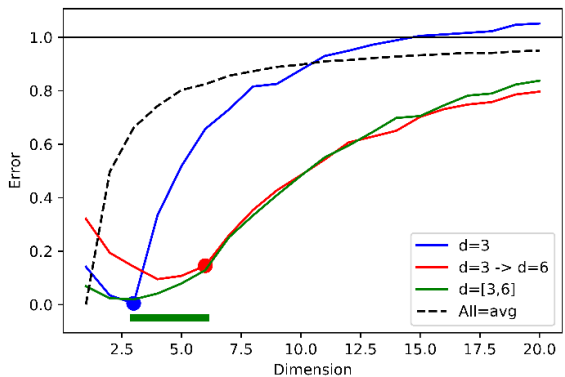

***Figure 7.*** This figure shows the performance of Badger when the number of dimensions of the optimization problem is changed from the dimension that it was trained on. The solid black line indicates chance level, whereas the dashed line indicates a solution that does not distinguish between the different dimensions of the problem (all dimensions are set to the mean value). When the model is trained on three dimensions (d=3), the best performance occurs there, but it still behaves better than chance level and better than the solution which does not distinguish between dimensions. A model trained on a range of dimensions from 3 to 6 sampled randomly generalizes quite well to lower dimension, and also demonstrates generalization up to the maximum of d=10 for this setup. The dots ● ● and horizontal bar ▬ correspond to training conditions associated with the different curves.

We believe that an MVE should evaluate an agent on its ability to adapt fast and efficiently. **Importantly**, an MVE should evaluate not only the agent's behavior and performance but also its internals (via white-box testing). For example, how are experts interacting, growing, the patterns of their communication, etc. Only observing their behavior externally might not be sufficient.

## Experimental Results

In trying to approach a general agent, we need a system which can generalize not just to different distributions of inputs and outputs or to different tasks, but to cases in which the inputs and outputs differ in format and in number. One stage of this would be to say that an agent trained on $N_{in}$ inputs and $N_{out}$ outputs should be able to generalize to $M_{in}$ inputs and $M_{out}$ outputs without re-training, where $N$ and $M$ can be different. A more severe requirement would be that an agent trained on image data should generalize to problems involving sound data or tabular data without re-training.

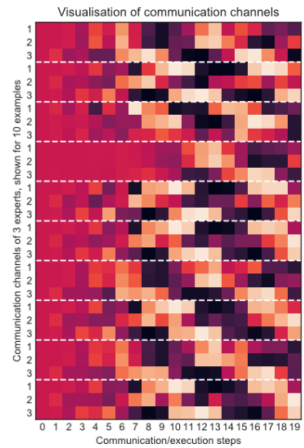

***Figure 8.*** Visualization of the dynamics of communication among the three experts comprising an agent in experiment shown in *Figure 6*.

We demonstrate that by structuring a network as a collection of experts with identical policies (internal weights), it is possible to train a method for function optimization that generalizes to different numbers of dimensions. This is done by allowing each expert to essentially address and 'push' values to different output dimensions by way of an attention mechanism (Vaswani et al. 2017) between the experts and a set of addresses associated with the different outputs. This way, if the number of outputs is changed, it simply means that there are more keys to which information can be pushed.

Using this sort of dynamically allocated key space can pose significant training difficulties, as initial policies tend to address all outputs simultaneously (as such, early local optima seen in training correspond to pushing the same value out on each output direction). However, longer training periods and training on a controlled cur-riculum of tasks can sometimes escape this local optimum of strategy and discover more general ways to search for the function optimum. In this case, we observe that the policy - if trained on both a small and large number of dimensions, can generalize to intermediate and unseen dimension counts, shown in *Figure 7*. While optimization weakens relative to what is possible as the number of dimensions is increased beyond the trained range, there is still a degree of generalization. The learned policy maintains a performance better than what one would see if it ignored the additional dimensions of the problem.

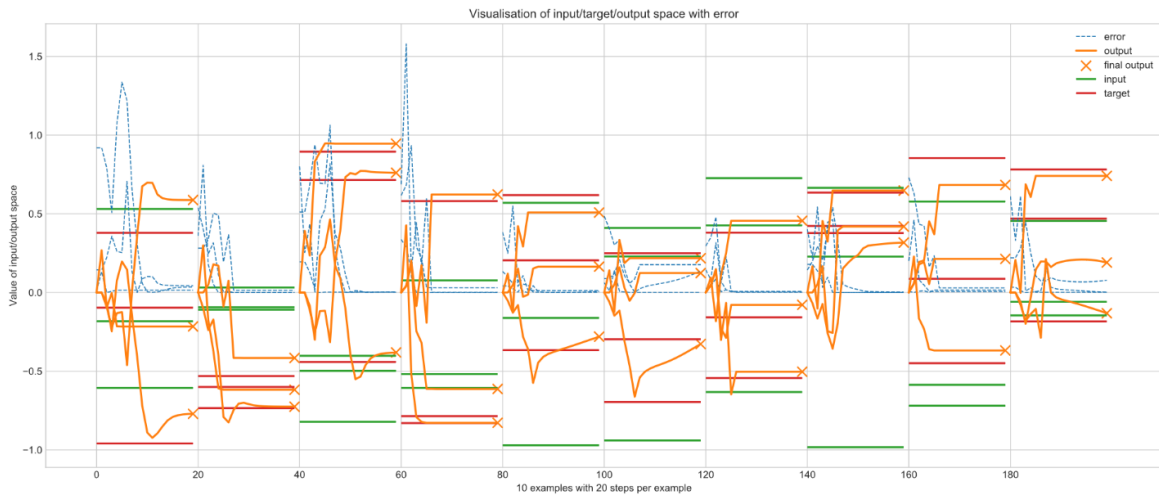

***Figure 9.*** In this experiment, the same setup as used in experiment shown in *Figure 6* is used. An agent comprises of three experts where only one expert receives the error information, which, collectively, the experts need to learn to minimize and hence guess the desired three dimensional output. Unlike in *Figure 6*, it is apparent that the agent found a significantly different strategy to arrive at the desired target. Rather than oscillating, the agent attempts to converge on the correct target as fast as possible and then stay there.

**Experiment – 'Guessing Game'**

In this experiment we are interested in analyzing how an expert policy can be found that allows an agent to learn to adapt based on an error signal provided on its input. The agent is provided with a predefined or variable number of inputs and outputs and an error signal that evaluates how closely the agent's outputs match the desired targets. This data is provided to the agent at every step of the inner loop. At a high level, one can imagine this task as "Guess *X* numbers" with feedback or as a learned optimization procedure.

**Hardwired Architecture -** *Figures 6* and *9* show results on this task under various hypotheses and with different learning objectives to show different aspects of the Badger architecture when the expert topology is fixed and expert connectivity is pre-defined. In *Figure 6*, the expert policy is trained to foster communication and to produce a target at the last step of the inner loop. In *Figure 9*, the conditions are similar, but here the agent is trained to converge on the desired target *as fast as possible*, rather than on the last step of the inner loop.

**Attentional Architecture -** In *Figure 7* the agent is tested on varying the number of dimensions of the optimization problem from the number of dimensions on which it was trained. In order to train this network to discover potential output dimensions to be optimized despite their addresses being random every time, we first train on a set of fixed addresses in d=3. This is done for 10k batches of size 50, changing the addresses only every 2k batches. Then, we switch to random addresses for every batch element and every batch and continue to train. The learning rate is lowered from 1e-4 to 5e-5 at 50k batches, and finally at 150k batches we record a checkpoint corresponding to the blue curve in *Figure 7*. We train at d=6 for an additional 50k batches to obtain the red curve, and then train for 100k batches with a random number of dimensions between [3,6] and a random number of experts in the range [5,40] to obtain the green curve.

If the initial stage of overfitting to fixed addresses is skipped, the model becomes stuck in a local optimum corresponding roughly to the black dashed line behavior - it can discover the average value over all dimensions to output on all outputs, but cannot distinguish between the outputs. When the initial overfitting stage is not allowed to converge deeply, there is a second local optimum that training can seemingly become stuck at approximately a loss value of 0.3 (for d=3). Demonstration code for this particular experiment can be found at https://github.com/GoodAI/badger-2019.

In addition to the shown results, we also observe that on tasks of this nature, increasing the number of experts can make training easier. In a related task, we observe that asymptotic performance actually scales with the number of experts even as the number of experts becomes larger than seen during training. This suggests that this approach may benefit from the sort of scalability that we mentioned in the motivations.

## Comparison to Related Work

In the following section, we will outline the main differences between our Badger approach and related work or areas of research that might evoke similarities.

### Artificial Neural Networks (ANN)

There are many differences between Badger and ANNs, both conceptually and technically. The comparison here is primarily for completeness.

ANNs:

- Inputs have fixed interpretation/function after training
- Number of input/output channels is constant and cannot vary between tasks
- Static architecture
- Learning occurs between nodes (edges are learned)

Badger:

- Roles of experts assigned dynamically at inference time
- Generalizes to different numbers and types of inputs/outputs (e.g. train on a 3-dimensional task, test on a 5-dimensional version)
- Can be trained to handle variations in architecture over batches or even during inference.
- Learning is entirely contained on each node (inner loop state updates, outer loop internal policy updates)

- Architecture can scale dynamically allowing for more computational capacity and power

**Meta-Learning and Learning to Learn**

'Learning to learn', or the ability to use past experience to facilitate further learning, has been observed in humans and other primates (Harlow 1949). Known in machine learning as meta-learning (Schaul and Schmidhuber 2010; Yao and Schmidhuber 1999; Juergen Schmidhuber, Zhao, and Wiering 1996; Thrun and Pratt 1998; Botvinick et al. 2019), the topic has recently attracted increasing interest (Andrychowicz et al. 2016; Finn, Abbeel, and Levine 2017). A variety of approaches have been proposed, mainly gradient-based ones where task adaptation is accomplished using gradient methods (Andrychowicz et al. 2016; Finn, Abbeel, and Levine 2017; Li and Malik 2017; Wichrowska et al. 2017) and memory-based ones where a learning procedure is acquired by for example a recurrent neural network (Ortega et al. 2019; Santoro et al. 2016; J. X. Wang et al. 2016; Duan et al. 2017; Mishra et al. 2018; Denil et al. 2019). Badger is an example of the latter class of architectures, with additional requirements regarding multi-agent cooperation, communication and extensibility.

**Modular Meta-Learning**

Meta-learning has also been extended to the modular setting, where different modules are used for specialization and diversification. Unlike in Badger, however, most works are limited to supervised learning and having different policies for each module (Alet et al. 2018; 2019; Alet, Lozano-Pérez, and Kaelbling 2018; Battaglia et al. 2018).

**Multi-Agent Reinforcement Learning**

The field of Multi-Agent Reinforcement Learning (MARL) deals with Reinforcement Learning problems where more than a single agent is active in an environment. Thorough recent reviews of this area can be found in (Hernandez-Leal, Kartal, and Taylor 2018).

- Badger experts are inside the agent, and they can interact with the environment only via an intermediary (membrane)

- Badger puts more emphasis on expert-to-expert communication (channels, topology, language), whereas in MARL, communication and language is optional (actions may be sufficient)

- All Badger experts aim to maximize a shared goal (agent's goal), whereas in MARL, shared goal for agents is optional

- All experts have the same expert policy, whereas in MARL this is optional

- Expert dynamics need not be driven by a reward function, but could learn to make use of other forms of feedback.

**MARL & Emergent Communication**

This sub-area of MARL focuses on how multiple agents can learn to communicate among themselves in order to better solve problems, or solve tasks that cannot be solved individually.

While early papers focused purely on benefits of explicit communication between agents (Sukhbaatar, Szlam, and Fergus 2016), more recent work focuses on specific properties of the communication/language that might be beneficial for some purposes. Examples of these properties can be e.g. interpretability of language (Mordatch and Abbeel 2018), or scalability of communication via local interactions (Jiang and Lu 2018) or targeted communication (Das et al. 2018).

Other important difference between these works is also in the assumption about the communication channel. While some works use differentiable communication channels, others focus on the more difficult non-differentiable communication scenario (i.e. communication through the environment)(Lowe et al. 2017).

Relevant requirements for the Badger architecture are mostly in shared policies, scalability of the communication (therefore decentralized and local properties) and the focus on meta-learning. (Jiang and Lu 2018) share some of concepts with Badger, however, our focus is on fast adaptation in the meta-learning setting, which is not common in the MARL field.

**Multi-Agent Meta-Reinforcement Learning**

Existing work on MARL doesn't yet focus on meta-learning. One exception is the work (Kirsch, van Steenkiste, and Schmidhuber 2019), which considers multiple agents, to learn a more universal loss function across multiple environments, where each agent is placed in a different environment. Compared to Badger, this work uses a standard MARL setting, where multiple agents are placed in an environment, rather than inside an agent. The agents also don't communicate.

**Neural Architecture Search**

This field is concerned with finding the optimal or better fixed topologies than ones designed by hand. Example works include (Cases et al. 2019; Castillo-Bolado, Guerra-Artal, and Hernandez-Tejera 2019) for dynamic topologies.

- Badger is not only concerned with finding a single topology, but rather with learning to dynamically evolve the topology of experts to facilitate fast adaptation to learning to solve new and unseen tasks

- Badger is a more universal decision-making system that contains aspects of neural architecture search

# Discussion

*Figure 3* on the first page shows our roadmap for Badger. It puts into perspective where we currently are, what our approximate plan is and how Badger is expected to be developed. Below, we briefly outline our current state of research and in the Appendix, we also outline our next steps, followed by a self-reflective analysis and criticism of this proposal, ending with some further notes on the topic and an FAQ, which we found to be an efficient source of information for many readers in the past.

**Current State**

Research of Badger architecture is still in its infancy. What we have achieved so far are encouraging results supporting some of the emergent properties of the proposed system:

- Evidence that one shared expert policy used by many experts, each having its unique internal state, can lead to learning/adaptation during the inner loop (learning while the expert policy is fixed = i.e. policy weights don't change)

- Evidence of one-shot learning

- Evidence of extensibility – adding more experts can improve agent's capacity on selected tasks (e.g. finding solutions faster or with better accuracy)

- Evidence of generalization across variable number of inputs

All these prototypes were realized on simplified toy-tasks. Extra effort would be needed to scale them up to a real world setting.

Also, the above prototypes are far from our final goal: *an expert policy with near zero task-specific properties, very general, coordinating experts to learn to solve new tasks.*

Expert policies that were trained in our current experiments are most likely coding very task-specific properties, thus not being sufficiently general, and will not scale outside of the training tasks.

# APPENDIX

## Next Steps

The following is a non-exhaustive list of what is ahead of us:

- Evidence of gradual learning (overcoming forgetting, skill reuse, skill reuse in order to learn new skills)
- Evidence of scalability to more complex problems, more cognitive skills, more intelligence
- Evidence of representation learning
- Evidence that tasks are not solved by the policy of a single expert directly, but by a pattern of activations / changes of internal memory states among experts
- Evidence of recursive self-improvement (or how to test whether the results of expert's communication is an improvement of learning capabilities for the agent)
- Evidence of 'intrinsic motivation' in the expert policy
- Evidence of 'skills cloning'
- Measuring the complexity of a Badger agent and inter-expert communication
- Explore the benefits of more than one shared expert policy
- Research the benefits of having more than one time-step inside the inner loop
- Research networks of thousands of interacting experts (until now, our prototypes scaled to ~100 experts)
- Benchmark Badger on standardized AI datasets (e.g. online learning MNIST, Omniglot, etc)
- Handcrafted / manual expert policy (as opposed to learned via an outer-loop)
- Evidence that 'Badger principles' can lead to scalable and incremental R&D that can take us to human-level AGI. In other words, how to prove that we need only 'good outer loop framework' and 'good tasks', in order to achieve AGI?
- Experimenting with different types of inter-expert messages, differentiable communications channels, discretized messages (vocabularies), topologies of experts, nested structures, being able to interpret the learned expert policy, and many more.

To accelerate progress on Badger research, an expertise in the following fields should be beneficial:

- Multi-agent reinforcement learning (MARL)
- Agent communication
- Emergence of language
- Auto-curricula
- Online meta-learning
- Others...

## Analysis & Criticism

*Some preliminary notes*: Converging to a proper expert policy may never happen. We will need to develop an MVE that provokes and evaluates the right kind of generality. As an example, take finding the right policy in (Baker et al. 2019). This might not have been possible a few years ago due to the lack of necessary compute.

### Criticism

- To learn a new type of building block – an expert – could be extremely time consuming and unattainable. The search space may be too large, non-linear and non-convex. The building block will be an expert, which itself is a neural network, and something that's an order of magnitude more computationally demanding than, for example, an individual neuron.

  - *Counter-argument*: If the opposite approach is to have one large neural network and one expects that a dynamic modularity similar to Badger will emerge in it, then this might be computationally even more infeasible

- The need to have the outer-loop practically multiplies the computation requirements as many times as there are inner-loop and communication steps (this could easily be 100-1000x)

  - *Counter-argument*: This is common to all meta-learning architectures

  - *Counter-argument*: If one can pay 1000x the cost once, but accelerate the future training of everyone who ever trains a model by 5%, that's still going to result in significant savings

- It may prove impossible to merge specialized expert policies - for example, an independently learned expert for gradual learning and an independently learned expert for image recognition, unable to be merged into a general coordination policy

- How is Badger different from let's say a Convolutional Neural Network (CNN)?

  - Similarities

    - They both share weights in various modules (each module has its own activations).

  - Differences:

    - CNN usually learns via gradient descent, by changing the kernel weights

    - Badger learns via activations, inside the inner loop, weights are fixed

    - CNN is traditionally one-directional, whereas the flow of signal between experts in Badger will be omni-directional and recurrent

    - The basic building block of a CNN is a neuron. The basic building block of Badger is an expert (a neural network)

    - Badger experts could communicate complex messages, feedback, memories, etc.

    - Badger agent can grow new experts

    - Badger's objective is fast adaptation to novel tasks, which is not intrinsic to CNNs

## Further Notes

- During the outer-loop and inner-loop, we may use gradient based learning, even differentiable messages between experts. However, in the deployment stage (after the expert policy has been fixed), we don't rely on gradient-based learning, nor differentiable communication explicitly. All learning in deployment stage is the result of inter-expert communication and changes in their internal states. Implicitly, a gradient-like procedure might occur as part of this process, but this is learned

- We expect that at the end of the outer-loop stage, expert policy will mainly contain a general coordination policy

with some form of learned credit assignment, with only very little task-specificity

- Experts are assumed to not have a spatial position within an agent or understanding of a space in this sense. Experts are just nodes in a graph, where edges do not need to represent a distance. But we are not ruling out the possibility of designing experts in a way which may be interpreted as if they were entities in an N-dimensional space, with distance between experts having a meaning. The learning of the expert policy in the outer loop would then have to discover an expert policy that can effectively use this kind of information. Our assumption is that this could lead to a more 'fuzzy' and 'less symbolic' type of communication between experts, which may be easier to learn for gradient-based learning systems in the outer loop.

- Learned strategies that are expected to emerge from the communication between experts are going to encode learning strategies, meta-learning strategies, and are going to influence the communication itself. It will be hard to tell them apart, hard to measure what communication pattern belongs to learning, meta-learning, or is just a simple skill.

- There are a few distinct objectives

  o *Outer loop* – aiming to discover an expert policy that leads to fast adaptation during the inner loop (time to solve, time to learn to solve)

  o *Inner loop* – adaptive performance on tasks

  o *Deployment* – is practically an inner loop running ad infinitum, expert policy doesn't change. The signal that may steer the agent's behaviour and learning is communicated through the same channels as during the inner loop.

- Tasks are not solved within the policy of a single expert. In other words, skills are not stored/represented in an individual expert.

- On the contrary, tasks are solved by a pattern of activations among experts. In other words, skills emerge from interactions of experts

- Actions are also a way of communicating

- 'Everything can be seen a communication problem'

- Language itself is a complex adaptive system (Beckner et al. 2009)

## FAQ

**Q: Why to have only one expert policy? (Or a few, but less than the number of experts)?**

**A:** By constraining the system to have only one shared policy used by all experts, we are pushing the learning system during the outer loop to discover a universal communication / coordination expert policy. It should contain as little task-specific elements as possible, because the capacity of one expert is not sufficient to encode all tasks. Furthermore, this constraint pushes the learning of task specific policies to be a result of interaction among experts during the inner loop. In other words, we want to force the learning system to discover the task policies as part of the inner loop, not as part of the outer loop. On the other hand, if we allowed each expert to have its own unique policy (number of expert policies would be equal to number of experts), the learning during the outer loop would most likely distribute the task policies into the expert policies, because the capacity of this network of networks would have enough capacity to store task specific policies in it.

**Q: What about hierarchical structures? Aren't we ignoring them?**

No. Even though these are not explicitly there, they can emerge in the internal dynamics of the experts communicating with each other.

## Acknowledgement

First we saw the video 'Honey Badger Don't Care' (Czg123 2011) which led us to seeing the documentary 'Honey Badgers: Masters of Mayhem' (BBC 2014). This tempted us to visit Stoffel, a famous honey badger living in Moholoholo, South Africa (Rosa 2019).

Honey badgers are an inspiring example of scalable intelligence and life-long learning. Much like our desired agent, they are able to come up with creative strategies and solve tasks way beyond their typical environment, thanks to general problem-solving, planning and fast adaptation abilities.

Honey badgers are also known for their grit, tenaciousness and 'can-do' attitude, which we believe are indispensable qualities for any team set to achieve 'impossible' goals.

For these reasons we couldn't help but fall in love with Honey Badgers and to aspire to their creative thinking, problem-solving and determination.

## EXTENDED BIBLIOGRAPHY

**Process Flow - Training Procedure**

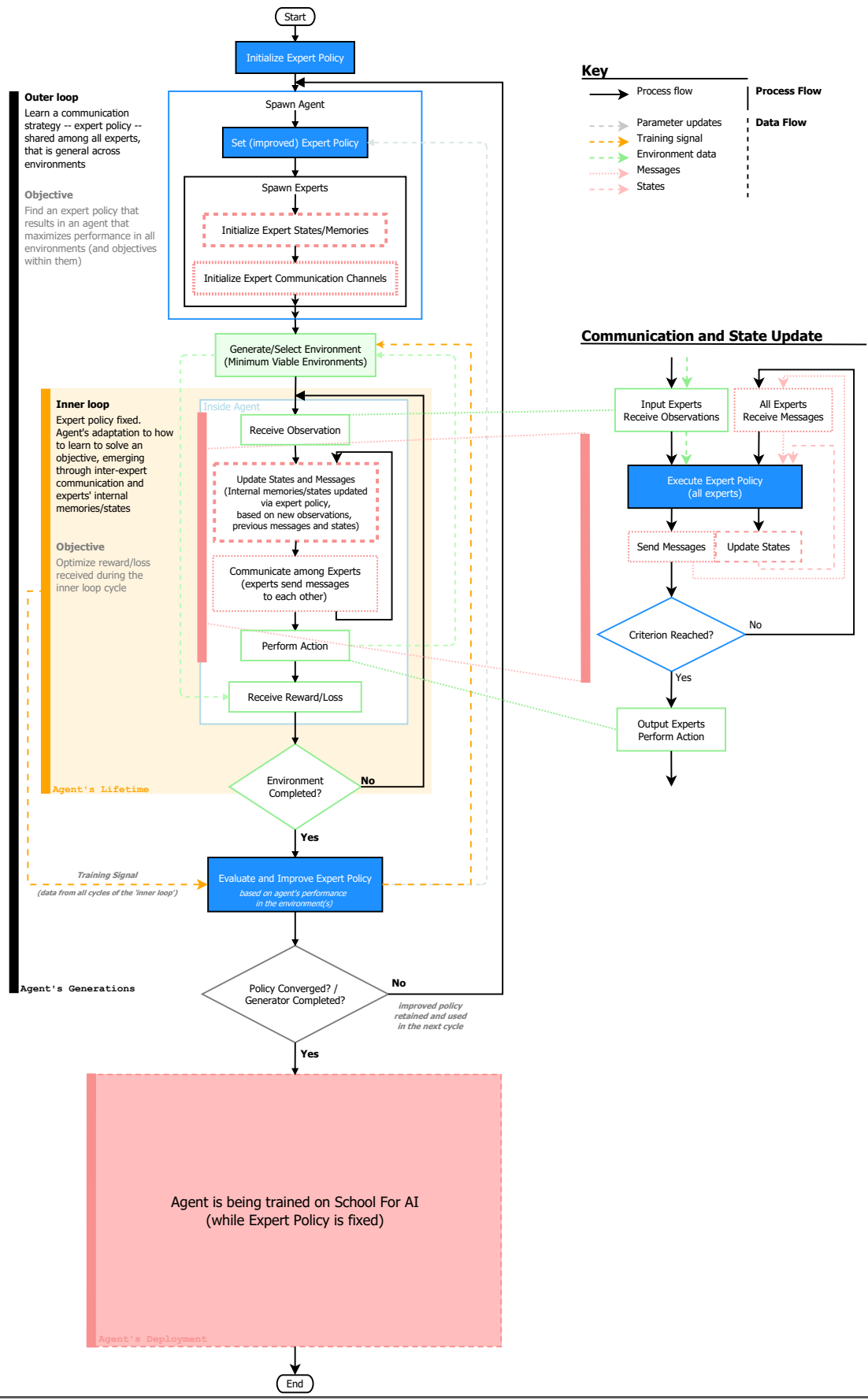